\documentclass{article}
\usepackage{hyperref}
\usepackage{graphicx} %
\usepackage{makecell}
\usepackage{url}
\usepackage{bbm}
\usepackage{amsmath}
\usepackage[numbers,sort]{natbib}
\usepackage[accepted]{icml2025}
\usepackage{times} 
\usepackage{multirow,multicol}
\usepackage{amsfonts}
\newtheorem{definition}{Definition}
\mathchardef\mhyphen="2D

\begin{document}

\icmltitlerunning{Sequence-Level Leakage Risk of Training Data in LLMs}
\twocolumn[
\icmltitle{Sequence-Level Leakage Risk of Training Data in Large Language Models} 
\author{Trishita Tiwari, G. Edward Suh}
\begin{icmlauthorlist}
\icmlauthor{Trishita Tiwari}{yyy}
\icmlauthor{G. Edward Suh}{yyy,comp}
\icmlcorrespondingauthor{Trishita Tiwari}{tt544@cornell.edu}
\end{icmlauthorlist}

\icmlaffiliation{yyy}{Cornell University}
\icmlaffiliation{comp}{Nvidia}
\vskip 0.3in
]
\printAffiliationsAndNotice{}

\begin{abstract}
    This work quantifies the risk of training data leakage from LLMs (Large Language Models) using sequence-level probabilities. Computing extraction probabilities for individual sequences provides finer-grained information than has been studied in prior benchmarking work. We re-analyze the effects of decoding schemes, model sizes, prefix lengths, partial sequence leakages, and token positions to uncover new insights that were not possible in previous works due to their choice of metrics. We perform this study on two pre-trained models, Llama and OPT, trained on the Common Crawl and The Pile respectively. We discover that 1) Extraction Rate, the predominant metric used in prior quantification work, underestimates the threat of leakage of training data in randomized LLMs by as much as $2.14$X. 2) Although on \emph{average}, larger models and longer prefixes can extract more data, this is not true for a substantial portion of individual sequences. $30.4-41.5\%$ of our sequences are easier to extract with either shorter prefixes or smaller models. 3) Contrary to previous beliefs, partial leakage in commonly used decoding schemes like top-$k$ and top-$p$ is \emph{not easier} than leaking verbatim training data. The aim of this work is to encourage the adoption of this metric for future work on quantification of training data extraction. 
\end{abstract}




\section{Introduction}
 Starting with ~\cite{carlini2021extracting}, several studies have documented leakage of training data from Large Language Models (LLMs). Many of these compare the extraction of training data across model sizes, different decoding schemes, training data duplication, training phases~\cite{carlini2022quantifying,biderman2024emergent,yu2023bag}, etc., and use a wide range of metrics for said comparisons. However, there are two major limitations in these prior quantification works, a) lack of analysis of LLMs with randomized output generation, and b) lack of probability driven per-sequence analysis, both of which we address in our work. We use per-sequence probabilities (defined in Section~\ref{sec:metric}) to calculate the risk of leaking a target suffix given its prefix in the training data. We evaluate this metric in various settings of training data leakage, such as: 
 \begin{enumerate}
     \item \textbf{Decoding Schemes}: Well-known prior work~\cite{carlini2022quantifying} claims that greedy decoding is the best decoding scheme to measure extraction. However, we show that this happens because ``extraction rate'', the metric used to quantify data leakage, consistently underestimates the risk of leakage from randomized decoding schemes such as top-$k$ and top-$p$. We observe that this underestimation can be as high as 2.14X for even a modestly strong adversary that can prompt an LLM $30$ times for each sequence to be extracted.

     \item \textbf{Partial Extraction}: Numerous prior works attempt to relax the constraints on the extraction of verbatim sequences from the training corpus by also allowing partial matches~\cite{yu2023bag,kassem2024alpaca,ippolito2023preventing,biderman2024emergent}. However, when extracting Named Entities, we discover that an overwhelming majority ($85.25-94.82\%$) of targets are \emph{not easier} to extract partially than exactly. This statistic is from Top-$k$ decoding ($k=5$), which~\cite{yu2023bag} found to perform the best for partial extraction. This forces us to think of better ways to capture and quantify partial leakage of potentially sensitive information from LLMs.

     \item \textbf{Model Sizes}: While ~\cite{carlini2022quantifying} highlights that on \emph{average}, there is a log-linear relationship between model size and extraction rates, we discover that individual sequences follow $6$ different types of trends, where a substantial percentage of samples ($30.4-31.7\%$) are \emph{easier} to extract with smaller models.

     \item \textbf{Prefix Lengths}: Similar to model sizes, prior work has identified a log-linear relationship between prefix lengths and extraction rates~\cite{carlini2022quantifying}. However, here too we find that individual samples follow $6$ different types of trends, where a substantial percentage of samples (32.7-41.5\%) are \emph{easier} to extract with shorter prefixes 
     
     \item \textbf{Token Position}: \cite{carlini2021extracting} posited that given a target sequence, extracting later tokens in the sequence might be easier than extracting earlier ones. Although this claim was based on intuition until now, we quantitatively verify that extracting later tokens in a target sequence is indeed upto $10.12$X easier than earlier tokens.
 \end{enumerate}

These findings provide new information that urges us to rethink some of our previous notions about memorization in LLMs.

\subsection{Roadmap}
The remainder of this paper is organized as follows: in Section~\ref{sec:motivation}, we outline the current research gaps to motivate the need for our work. In Section~\ref{sec:background}, we discuss background concepts. Section~\ref{sec:metric} includes the definitions of the metrics we use for our analysis. We present our analysis in Section~\ref{sec:eval} and highlight any novel conclusions. Section~\ref{sec:rel} outlines related literature, and we conclude this work in Section~\ref{sec:conclusion}

\section{Motivation: Issues with existing measure of leakage}
\label{sec:motivation}

We discuss the two main research gaps in existing work to motivate this study.

\subsection{Probabilistic Quantification of Training Data Leakage} 

\cite{carlini2022quantifying}, one of the first works to attempt to quantify training data leakage in LLMs, defines memorization as the following: 
\begin{definition}
Memorization: A string s is extractable with $k$ tokens of context from a model $f$ if there exists a
(length-$k$) string $p$, such that the concatenation $[p || s]$ is contained in the training data for $f$, and $f$
produces $s$ when prompted with $p$ using greedy decoding.  
\end{definition}
Thus, by definition, the traditional notion of memorization only includes non-randomized generation schemes such as greedy decoding. This makes the definition deterministic, making memorization easy to quantify. However, a major disadvantage is that fixing the decoding scheme excludes samples that would otherwise leak through other (randomized) decoding schemes. In general, randomized generation implies that the same prompt can lead to different output strings. These strings are generated in a probabilistic fashion, and random decoding is commonly used in generative models deployed in production~\cite{chatgpt,claude,cohere,mistral}. Thus, from a practical perspective, analyzing training data leakage risk through a probabilistic lens is more aligned with real-world settings.  Furthermore, leakage probabilities capture more than just a binary label of ``extracted'' or ``not extracted'', providing more fine-grained information about the target sequences.


\subsection{Sequence-Level Probabilities}
Prior works on quantifying training data leakage typically only measure population-level metrics (e.g extraction rates, precision, recall, average hamming distance, etc.)~\cite{carlini2021extracting,carlini2022quantifying,yu2023bag,kassem2024alpaca,lukas2023analyzing,huang2022large}. They do so by testing a model with a large number of sequences from its training data, and then compute an averaged extraction metric over this population of sequences. Although these metrics are useful in certain contexts, they offer limited insights as they condense all the information to a singular statistic. While this could be appropriate in deterministic models where each training sequence can only be ``extracted'' or ``not extracted'', this is not the case in a probabilistic setting. In such settings, it would also be useful to get the probability of leaking each sequence when the model is prompted many times. Firstly, this allows us to provide guarantees on the risk of leaking each individual sequence. This can be important in the case where the individual sequences in the training data consist of, say, user information or copyrighted content. Secondly, it
allows us to study the effects of different models, hyper-parameters, etc., on the entire distribution
of training sequences, rather than their effect on a handful of averaged statistics from the distribution. Indeed, depending on the use case of the model, different aspects of the distribution would be
more or less important, and we wish to make it easier for ML engineers to make informed decisions based on more fine-grained data, and not just the average. Finally, the averaged statistics in prior work~\cite{carlini2022quantifying,yu2023bag} assume
that any adversary trying to extract certain training data can only prompt the model \emph{once}
per sequence. This is a gross underestimation of an attackers capabilities as most production models deployed allow multiple, if not an unrestricted number, of queries for each user.

\section{Background}
Here, we provide an overview on the language modeling objective, logit normalization schemes, and decoding schemes.

\label{sec:background}
\subsection{Language Modeling Objective}
The language modeling objective consists of the task of a model $M$ parameterized by $\alpha$ to predict the next token when given a sequence of tokens $t_{1:j}$. The model does so by outputting a probability distribution over the next token $t_{j+1}$, like so:

\begin{equation*}
    P(t_{j+1} | t_{1:j}) = M_\alpha(t_{1:j})
\end{equation*}

The model is then fed tokens in an auto-regressive manner, where a token from the output distribution is sampled and appended to the token sequence, after which the process is repeated to generate the next token. 

More formally, the model $M_\alpha =  d_\phi \circ  n_\psi \circ m_{\mathcal{V},\theta}$ where $\alpha = \{ \mathcal{V}, \theta, \psi, \phi \}$. $m_{\mathcal{V},\theta}$ represents the forward pass of the model, which has a vocabulary $\mathcal{V}$ and is parameterized by $\theta$. $d_\phi$ and $n_\psi$ represent the \emph{decoding} and \emph{normalization} schemes, both of which are discussed below.

\subsection{Normalization}
After the forward pass, a normalization function $n_\psi$ converts the model's output logit vector to a probability distribution. We now describe a popular normalization function and its variant:
\begin{enumerate}
    \item \textbf{Softmax} Each un-normalized logit $z_j$ in a vector of $N$ logits is normalized through the following equation:
    \begin{equation*}
        \sigma(z_j) = \frac{e^{z_j}}{\sum_{j=0}^Ne^{z_j}}
    \end{equation*} 
    \item \textbf{Softmax with Temperature} Softmax can also be used with a ``temperature'', which involves dividing the logits of the model by a chosen constant $K$ (the temperature). This causes the shape of the resulting distribution after normalization to be either ``sharper'' or ``flatter'' depending on the value of $K$. This modification of the distribution shape allows for more or less diversity in the sampled tokens depending on $K$. Below is what applying a temperature to softmax looks like:   
\begin{equation*}
    \sigma(\frac{z_j}{K}) = \frac{e^{\frac{z_j}{K}}}{\sum_{j=0}^Ne^{\frac{z_j}{K}}}
\end{equation*}
\end{enumerate}

\subsection{Decoding Schemes}
After normalization, the output distribution goes through a decoding scheme $d_\phi$. $d_\phi$ modifies the distribution and samples one token from the modified distribution. There are several different decoding schemes used; we discuss some popular ones below:
\begin{enumerate}
    \item \textbf{Greedy} Greedy decoding involves selecting the token with the highest probability from the output distribution. This leads to deterministic output--the same prompt to the model will always yield the same distribution, which will lead to the same token being chosen.
    \item \textbf{Random Sampling} Random sampling simply samples a token out of the distribution of tokens without further modifying it. This leads to non-deterministic output since different tokens can be sampled in different runs. 
    \item \textbf{Top-\textbf{\em k}} Like random sampling, top-$k$ is also a probabilistic decoding scheme. However, instead of sampling from the entire distribution of tokens (which generally consists of tens of thousands of tokens), we limit the sampling window to the $k$ tokens with the highest probabilities. This truncated distribution is then renormalized before a token is sampled from it.
    
    \item \textbf{Top-\textbf{\em p}} Like top-$k$, Top-$p$ involves sampling from a truncated distribution; however the distribution only includes tokens with the highest likelihood whose sum is $<= p$. Like top-$k$, this truncated distribution is also renormalized before a token is sampled from it.
\end{enumerate}

\section{Metric Definitions}
\label{sec:metric}

In this section, we describe the metrics that we use for our analysis in Section~\ref{sec:eval}. For the sake of brevity, we also define abbreviations for each of these metrics to use in subsequent sections.

\paragraph{Preliminaries}
Like in Section~\ref{sec:background}, we use a model $m_{\mathcal{V},\theta}$. $m_{\mathcal{V},\theta}$ has a vocabulary $\mathcal{V}$ and is parameterized by $\theta$. Its output logit vector is normalized into a probability distribution by a normalization function $n_\psi$ (e.g., softmax). After this, a token is sampled via a decoding function $d_\phi$ (e.g., top-$k$, top-$p$, etc). For brevity, we define the following composition of the above functions $M_\alpha = d_\phi \circ n_\psi \circ m_{\mathcal{V},\theta}$ where $\alpha = \{ \mathcal{V}, \theta, \psi, \phi \}$. $M_{\alpha}$ thus represents a model $m_{\mathcal{V},\theta}$ with a fixed normalization function $n_\psi$ and decoding scheme $d_\phi$.

\begin{definition}
Token Probability: The probability of producing a token $t_{j+1}$ when the model is prompted with a sequence of $j$ tokens $t_{1:j}$ is defined as by \emph{Token Probability (TP)}:
\begin{align}
\begin{split}
    TP(t_{j+1},t_{1:j},M_\alpha) &= P(t_{j+1}|t_{1:j};M_\alpha) \\ &= M_\alpha(t_{1:j})[t_{j+1}]
    \label{eq:tlp}
\end{split}
\end{align}
    
    
\end{definition} 

\begin{definition} Exact Sample Probability: The probability of producing a sequence of $m$ tokens $t_{j+1:j+m}$ when the model is prompted with a $j$-token prefix $t_{1:j}$, is the Exact Sample Probability (ESP) defined as:
\begin{equation}
    ESP(t_{j+1:j+m},t_{1:j},M_\alpha) 
    = \prod_{n=j+1}^{j+m}TP(t_n,t_{1:n-1},M_\alpha)
\label{eq:slp}
\end{equation}
\end{definition}

Equation~\ref{eq:slp} is a closed form equation for the probability of leaking our target sequence verbatim. However, in practice, one might also want to relax this requirement and measure the risk of leaking parts of the target sequence--i.e., by allowing a subset of $n$ tokens within the length-$m$ target sequence to be incorrectly extracted. 

\begin{definition} 
n-Inexact Sample Probability: We let $\mathcal{E}(t_{1:m},n)$ define the set of all token sequences of length $m$ that are an edit distance of $n$ from $t_{1:m}$. We measure the leakage probability of an $m$-token suffix $t_{j+1:j+m}$ allowing $n$ tokens to be incorrectly matched when the model is prompted with a $j$-token prefix $t_{1:j}$, through the $n$-Inexact Sample Probability (n-ISP):
\begin{align}
    \begin{split}
    n\mhyphen ISP(n,t_{j+1:j+m},t_{1:j},M_\alpha) \\= \sum_{g \in \mathcal{E}(t_{j+1:j+m},n)}ESP(g,t_{1:j},M_\alpha)
    \label{eq:inex}    
    \end{split}
\end{align}
\end{definition}

\begin{figure}
    \centering
    \includegraphics[trim={0.1in 0 0 0},clip]{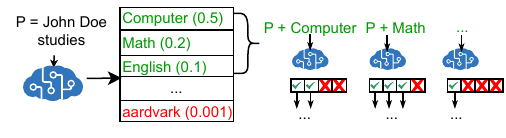}
    \vspace{-0.3in}
    \caption{Approximating Inexact Sample Probabilities (Equation~\ref{eq:inex-approx}) by only iterating over tokens in the head of the distribution. An upper bound on the error term, $\varepsilon$, can be computed by summing the probabilities in the ignored tails (see Appendix~\ref{app:isp}).}
    \label{fig:islp-approx}
\end{figure}

Computing Equation~\ref{eq:inex} can become expensive as the set $\mathcal{E}(t_{j+1:j+m},n)$ can become fairly large. For example, allowing $2$ from $4$ tokens in the target suffix to be extracted incorrectly would involve summing the probabilities over $\mathcal{E}(t_{j+1:j+4},2) = \binom{4}{2}\mathcal{V}^2$ strings, where $\mathcal{V} = 32,000$ for Llama. Furthermore, as $n$, $m$, and $\mathcal{V}$ increase, this figure will increase very quickly and become infeasible to compute. To mitigate this, we can approximate Equation~\ref{eq:inex} by only iterating on a subset of the highest-probability sequences $S \subset \mathcal{E}(t_{j+1:j+m},n)$, as shown in Figure~\ref{fig:islp-approx}. 


\begin{definition}
Approximate n-Inexact Sample Probability: Let $S \subset \mathcal{E}(t_{j+1:j+m},n)$. n-ISP can thus be approximated as follows:
\begin{align}
\begin{split}
    n\mhyphen ISP(n,t_{j+1:j+m},t_{1:j},M,\theta) \\= \sum_{g \in \mathcal{S}}ESP(g,t_{1:j},M,\theta) + \varepsilon
    \label{eq:inex-approx}    
\end{split}
\end{align}
\end{definition}

\subparagraph{Error Approximation ($\varepsilon$)} We can also estimate an upper bound on $\varepsilon$ by summing the worst case probabilities of the sequences that were not included in $S$. This is described in more detail in the Appendix~\ref{app:isp}.



\section{Evaluation}
\label{sec:eval}
\subsection{Evaluation Setup}
We evaluate these metrics on $2$ recent generative models: Llama and OPT, which were pretrained on the Common Crawl (Llama), or The Pile (OPT). We perform an evaluation of leakage risk in $3$ different settings: 
\begin{enumerate}
    \item Following~\cite{lukas2023analyzing}, we randomly sample $10,000$ sequences that contain Named Entities (NEs) from each of the datasets. We do this since NEs can simulate sensitive data that might be targets for extraction since they are typically names of individuals, organizations, places, etc. We must note that this is simply to emulate a realistic use-case, but in reality, model developers would repeat the measurements with sequences that they consider sensitive. For consistency, each of our sequences consists of a prefix of $50$ tokens followed by a $4$-token long Named Entity. 
    \item We also simulate extraction of longer sequences, which could be useful to measure the risk of leaking copyrighted data. For this, we randomly sample $10,000$ sequences with $50$-token prefixes and $50$-token suffixes from each dataset, similar to prior work~\cite{biderman2024emergent,carlini2022quantifying}, and,
    \item Lastly, we also perform our analysis on same dataset from The Pile as~\cite{yu2023bag}, which consists of $1$-eidetic sequences as defined by the 2023 SATML training data extraction challenge. This dataset consists of $15,000$ sequences, each with a prefix of $50$ tokens followed by a suffix of $50$ tokens.
\end{enumerate}

    

\subsection{Impact of decoding Schemes}

\begin{figure*}
    \centering
    \includegraphics{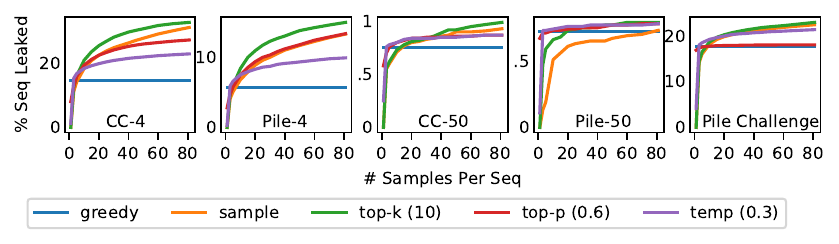}
    \vspace{-0.2in}
    \caption{The percentage of sequences that can be leaked for different datasets as the number of times the model is prompted per sequence increases.}
    \label{fig:decode}
    \vspace{-0.15in}
\end{figure*}

\paragraph{Experiment} In Figure~\ref{fig:decode}, we compare the percentage of our $5$ datasets that leak across $6$ different decoding schemes as the number of times we prompt the model for each sequence increases. We perform the experiment by counting the number of sequences that would be leaked at least once if a model was sampled $X$ times per sequence, as a function of increasing $X$. (This is equal to the number of samples whose ESP is greater than $\frac{1}{X}$ at every $X$). These decoding schemes are used with hyper-parameters that the most recent state-of-the-art work by~\cite{yu2023bag} found to perform best for training data extraction. 

\paragraph{Extraction Rate Underestimates Leakage in Randomized Decoding} We note that since greedy decoding is deterministic, its resulting trend in Figure~\ref{fig:decode} is simply a horizontal line equal to the \emph{extraction rate} \cite{carlini2022quantifying,yu2023bag,carlini2021extracting}. For the other randomized decoding schemes, the number of sequences leaked increases as we sample the model more. We see that it takes randomized decoding schemes between $3-81$ samples to outperform greedy decoding. In fact, all except one decoding scheme (``sample'' for PILE$-50$) outperform greedy after about $3-20$ samples. This means that if a targeted attacker samples each sequence even $20$ times, it is enough to extract more sequences than is possible with greedy decoding. However, as mentioned, comparisons in prior work like~\cite{carlini2022quantifying} and~\cite{yu2023bag} rely solely on extraction rate. Since extraction rate is simply an average of the ESP of suffixes, it tells us what percentage of all suffixes could be extracted if an attacker prompts the LLM with their respective prefixes \emph{once}. This means that the prior comparisons are based simply on the first data point of each graph in Figure~\ref{fig:decode}, where greedy decoding clearly outperforms all randomized decoding schemes. However, since the performance of randomized decoding increases drastically after just a few more samples, it highlights how using solely the first data point for comparison would underestimate the leakage risk for randomized decoding.

\subsection{Inexact Leakage}
\paragraph{Experiment} Since partial leakage, especially of sensitive information such as names and other PII, can be a concern, we compute the likelihood of such leakage. For this experiment, we compute ESP, 1-ISP and 2-ISP for our datasets (Common Crawl and the Pile) that simulate PII with 4-token Named Entities. We examine the probabilities of leaking individual sequences, the effects of allowing a different number of tokens to be incorrect, and the effects of different decoding schemes. We also closely examine the $100$ sequences with the highest ISP and provide our commentary on these sequences. We discuss our observations below. 

\paragraph{Partial leakage is \emph{not} more likely} Many works attempt to relax the constraints of verbatim training data leakage by also counting partial leakages~\cite{yu2023bag,biderman2024emergent,kassem2024alpaca}. The intuition is that not requiring all tokens to match might make the sequences more likely to leak. However, when computing $1$ or $2$-ISP with top-$k$ ($k=5$) decoding, we see that a majority ($85.25-94.82\%)$ of Named Entities do \emph{not} see an improvement in their likelihood of extraction even if we allow tokens to be mismatched. This is because of the \emph{conditional} nature of language modeling objective. Simply put, if we want to allow, say, the first token to be incorrect, then the model has to produce the remaining tokens correctly while being conditioned on an incorrect token. In the majority of cases, this is harder for the model to do, making partial matches harder to produce than exact matches. 

These results imply that we need to rethink how we quantify partial leakage, especially for potentially sensitive data like Named Entities. Since token-based partial matches are uncommon, perhaps a better approach to capture partial leakage would be to measure the semantic closeness between the training data and the model's generation, regardless of whether the tokens match or not.

\paragraph{Effect of number of incorrect tokens} We observe that as the number of tokens that we allow to be incorrect increases, the likelihood of observing a partial leakage also increases. For example, the number of sequences that are easier to extract when we allow $2$ out of $4$ tokens to be incorrect is $12.36-14.75\%$, compared to only $5.17-6.93\%$ when we allow just $1$ token to be incorrect. While this still suggests that it is for the most part easier to extract a sequence verbatim than partially, as the number of tokens allowed to be incorrect increases, the likelihood of observing partial matches also increases.

\paragraph{Effect of decoding schemes}
In Table~\ref{tab:partial-decode}, we compare $1$-ISP across $4$ decoding schemes: top-$k$ ($k=5$), top-$p$ ($p=0.6$), temperature sampling ($0.1$) and regular sampling. The parameters for each of these decoding schemes were empirically found to be optimal for partial extraction by~\cite{yu2023bag}.

We see that even though the $1$-ISP for regular sampling (Column $4$) is a lower bound\footnote{Since computing $1$-ISP for regular sampling is very computationally intensive, we compute a lower bound using the methodology described in Section~\ref{sec:metric}. Here, we only use the first $90\%$ of the probability mass instead of the full distribution at every step of the calculation.}, $36.2-51.9\%$ of sequences have a higher chance of partial extraction with $1$ incorrect token than with exact extraction. However, even though the corresponding numbers for top-$k$ and top-$p$ are exact values, they are much lower--only $9.26-11.0\%$ and $10.14-12.71\%$ respectively. This is despite the fact that top-$k$ ($k=10$) is noted as one of the best performing decoding schemes in~\cite{yu2023bag}. 

\begin{table}
    \small
    \centering
    \caption{Percentage of Named Entities that see an improvement in extraction probability if $1$ token is allowed to be incorrect (1-ISP), based on decoding scheme for different datasets. The percentages for Regular Sampling (Column $4$) is a lower bound, while the percentages in the other $3$ Columns are exact numbers.}
    \vspace{0.1in}
    \begin{tabular}{|c|c|c|c|c|}
        \hline
        \multicolumn{5}{|c|}{Samples easier to extract w/ single token mismatch} \\
        \hline
        \multirow{2}{*}{Dataset} & \multicolumn{4}{c|}{Decoding Scheme} \\ \cline{2-5}
         & \makecell{Top-$k$ \\ ($k=10$)} & \makecell{Top-$p$ \\ ($p=0.6$)} & \makecell{Temp \\ ($0.1$)} & \makecell{Regular \\ Sampling} \\ \hline
         CC & 10.95 & 10.14 & 26.42 & 36.2 \\ \hline
         Pile & 9.26 & 12.71 & 40.0 & 51.87 \\ \hline
    \end{tabular}
    \label{tab:partial-decode}
\end{table}

On examining breakdown of individual sequences closely, we note that the same sequence is more likely to be partially extracted through regular sampling than top-$k$ because regular sampling allows a significantly larger number of partially matched sequences to have a non-zero probability. These, when summed together (Equation~\ref{eq:inex}), can sometimes give us a larger ISP than extracting the sequence exactly. However, top-$k$ zeros out the probabilities of most sequences in the distribution, effectively allowing very few partially correct sequences to have non-zero probability.

We note that~\cite{yu2023bag} use only $5$ trials to compare partial extraction across different decoding schemes, which could explain why they found top-$k$ decoding to outperform regular sampling. $5$ trials is not enough to observe very low-probability partial sequences, which are dominant in regular sampling. This could lead to an underestimation of the likelihood of partial extraction in regular sampling compared to other decoding schemes. The metrics we use do not face this issue since they're closed-form, and thus, give us exact probabilities.  

\paragraph{Qualitative observations : sequences with high ISP}
While partial matches are, for the most part, less likely than exact matches, we closely examine partial sequences that do not fit this trend (i.e., those that are easier to produce than their exact counterparts). Looking at the $100$ sequences with the highest inexact probabilities, we see two patterns, both of which we discuss below.
\begin{enumerate}
    \item \textbf{Incorrect Trailing Tokens} There are respectively $4$ and $6$ ways for $1$ and $2$ tokens to be incorrect in a $4$-token sequence. However, in Table~\ref{tab:partial-combo}, we find that certain permutations have a much higher likelihood than others. In both $1$ and $2$-ISP we see that the most likely permutations are ones where either the last or last two tokens are incorrect. As explained above, this happens because all other permutations require the model to be able to produce subsequent tokens correctly while being conditioned on an incorrect token(s), and are thus far less likely. The upper $2$ rows of Table~\ref{tab:partial-seq} show examples of such sequences.  
    \item \textbf{In-Domain Token Substitutions} While most partial matches are likely to have incorrect trailing tokens, Table~\ref{tab:partial-combo} shows that a minority of sequences can also be generated with incorrect leading tokens. Upon examining these more closely, we observe that this only happens when there exists an incorrect token that has a semantic meaning similar to that of the original token in the sequence. This condition is necessary for the model to be able to correctly produce the other tokens after being conditioned on an incorrect token. Since this is a hard requirement to satisfy, it may explain why partial matches are not more likely than exact matches. The lower $2$ rows of Table~\ref{tab:partial-seq} show examples of such sequences. 
\end{enumerate}



\begin{table}[]
    \small
    \centering
    \caption{Out of the $100$ partial matches that have the highest probabilities, this table shows which incorrect permutations were the most likely to be generated for $4$-token Named Entities in the Common Crawl (CC) and Pile datasets.}
    \vspace{0.1in}
    \begin{tabular}{|c|c|c|c|c|c|}
    \hline
     \makecell{1 Token \\ Incorrect \\ Positions} & CC & Pile & \makecell{2 Tokens \\ Incorrect \\ Positions} & CC & Pile \\ \hline
     (1) & $18$  & $21$ & (1,2) & $5$ & $5$ \\ \hline
     (2) & $11$ & $2$ & (1,3) & $2$ & $11$ \\ \hline
     (3) & $11$ & $6$ & (1,4) & $8$ & $4$ \\ \hline
     (4) & $60$ & $71$ & (2,3) & $2$ & $1$ \\ \hline
     - & - & - & (2,4) & $2$ & $1$ \\ \hline
     - & - & - & (3,4) & $81$ & $78$ \\ \hline
    \end{tabular}
    \label{tab:partial-combo}
    \vspace{-0.05in}
\end{table}

\begin{table}
    \small
    \caption{Examples of Named Entities found in the Common Crawl dataset that were tested for their extraction probability using top-$k$ decoding. Whitespace is used to indicate the model's tokenization. The first column shows the correct entity as found in the dataset, and the second column shows a higher-probability partial match. Parentheses enclose the probabilities for each of these sequences.}
    \vspace{0.1in}
    \centering
    \begin{tabular}{|c|c|}
    \hline
    Ground Truth & Model Generation \\ \hline
    G im po \textbf{Airport} (0.032) & G im po \textbf{International} (0.92) \\ \hline
    Egg l estone \textbf{Abbey} (0) & Egg l estone \textbf{is} (0.4) \\ \hline
    Mar iano \textbf{Rah} oy (0) & Mar iano \textbf{Raj} oy (0.97) \\ \hline
    \textbf{Karl} Ho er ig (0) & \textbf{Stephen} Ho er ig (0.23) \\ \hline
    \end{tabular}
    \label{tab:partial-seq}
    \vspace{-0.1in}
\end{table}


\subsection{Effect of Model Size}
\begin{figure}
    \centering
    \includegraphics[trim={0.1in 0 0.1in 0},clip,width=0.35\textwidth]{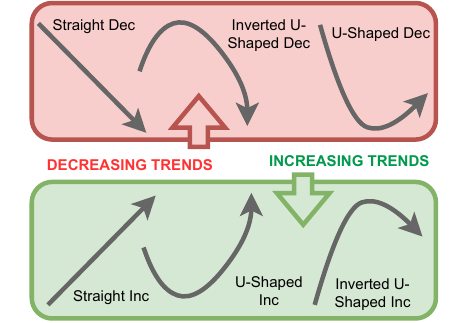}
    \caption{The $6$ types of trends observed in ESP of different sequences as the prefix length (or model size) increases. The following is a description of the trends: a) Straight-Dec: end point is the minimum point, no intermediary max/min points b) Inverted U-shaped Dec: end point $<$ start point, but an intermediary point is the maximum, c) U-Shaped Dec: end point $<$ start point, however an intermediary point is the minimum, d) Straight Inc: end point $>$ start point, no intermediary max/min points, e) U-Shaped Inc: end point $>$ start point, an intermediary point is the minimum, f) Inverted U-Shaped Inc: end point $>$ start point, but an intermediary point is the maximum}
    \vspace{-0.1in}
    \label{fig:trends}
\end{figure}

The current understanding based on~\cite{carlini2022quantifying} is that \emph{average} extraction rate increases log-linearly as the model-size increases. However, as discussed above, the extraction rate is averaged over an entire population of sequences. To obtain finer-grained information, we examine how the ESP of \emph{each} sequence changes as the number of parameters in OPT increases (125M, 350M, 1.3B, 6.7B, 13B). Here, we discover the $6$ trends (illustrated in Figure~\ref{fig:trends}) that the ESP of different samples can take as the model-size increases. The percentage of samples that demonstrate these different trends are shown in Table~\ref{tab:modelsize-trends}.

\paragraph{Discussion} In Table~\ref{tab:modelsize-trends}, we see that between $0.18-10.05\%$ (sum of first $3$ rows) of sequences become harder to extract as the size of the model increases. 
Upon examining these sequences closely, we notice that most of them appear to be unnatural-looking, out-of-distribution text for the $50$-token suffixes. For the $4$-token suffixes, there seems to be an equal proportion of natural and unnatural text that follow a trend of decreasing ESP. We provide examples of these sequences in Appendix~\ref{app:model-size-ex}. 

Curiously, $27.41-40.16\%$ of sequences are easier to extract with an intermediary model-size than the largest model (row $6$). Unlike sequences that have a decreasing ESP with increasing model-size, these sequences are predominantly natural-looking text. In the case of extraction of Named Entities, this trend is followed by the largest proportion ($40.16\%$) of sequences in the population, however for longer suffixes, this is no longer the majority trend.

This implies that smaller models might be perfectly good at leaking shorter extraction targets like Named Entities, and so care must be taken when training on sequences containing these.

\begin{table}
    \small
    \centering
    \caption{The percentage of samples whose ESPs follow the $6$ trends illustrated in Figure~\ref{fig:trends} as the model-size of OPT increases as follows: 125M, 350M, 1.3B, 6.7B, 13B.}
    \vspace{0.1in}
    \begin{tabular}{|c|c|c|c|}
         \hline
    
        \multirow{2}{*}{Trend} & \multicolumn{3}{c|}{Suffix Length} \\ \cline{2-4}
        
         & 4  tokens & \multicolumn{2}{c|}{50  tokens} \\ \hline

        \makecell{Straight Dec} & 0.64 & 0.07 & 0 \\ \hline

        \makecell{U-Shape Dec} & 4.84 & 0.7 & 0.073 \\ \hline
         
        \makecell{Inverted-U-Shaped Dec} & 4.57 & 0.36 & 0.11 \\ \hline
        
        \makecell{Straight Inc} & 35.51 & 61.96 & 67.94 \\ \hline

        \makecell{U-Shaped Inc} & 28.05 & 9.9 & 6.56 \\ \hline

        \makecell{Inverted-U-Shaped Inc} & 40.16 & 30.41 & 27.41 \\ \hline
        
    \end{tabular}
    \vspace{-0.2in}
    \label{tab:modelsize-trends}
\end{table}

\subsection{Effect of Prefix Length}
Similar to model sizes, the current understanding based on~\cite{carlini2022quantifying} is that the extraction rate increases log-linearly as the prefix length increases. To obtain a finer-grained insight, we examine how the ESP of \emph{each} sequence changes as the prefix length increases from $10$ tokens to $50$ tokens in increments of $10$. Here too, we discover $6$ different types of trends that the ESP of each sample can follow as the length of the prefix increases. The schematics of these different trends are shown in Figure~\ref{fig:trends}, and the number of samples demonstrating these different trends are outlined in Table~\ref{tab:prefix-trends}.

\paragraph{Discussion}
The sum of rows $1-3$ in Table~\ref{tab:prefix-trends} shows $17.58\%-18.15\%$ of Named Entities have an overall \emph{decreasing} trend, where the ESP decreases as the prefix length increases. This is surprising as a non-negligible amount of the population follows a trend that is opposite to that established in prior work~\cite{carlini2022quantifying}. However, when considering the extraction of longer sequences, this percentage drops to between $4.09-4.57\%$. While much a much smaller proportion, this still corresponds to $\approx400$ out of our $10$K tested samples.

The last $3$ rows in Table~\ref{tab:prefix-trends} consist of samples that are easier to extract as the prefix lengths increase. The final row shows samples that are easiest to extract by an intermediary prefix length, and not the longest length. When extracting Named Entities, we see that the majority of sequences ($33.14-41.45\%$) follow this trend. When extracting longer, $50$-token sequences, this trend is no longer the majority ($32.68-33.76\%$), but is still a substantial percentage of the population.

\begin{table}
    \small
    \centering
    \caption{The percentage of samples whose ESPs follow the $6$ trends illustrated in Figure~\ref{fig:trends} as the prefix length increases from 10 to 50 tokens in increments of 10 tokens}
    \vspace{0.1in}
    \begin{tabular}{|c|c|c|c|c|c|}
         \hline

        \multirow{2}{*}{Trend} & \multicolumn{5}{c|}{Suffix Length} \\ \cline{2-6}
         
        & \multicolumn{2}{c|}{4  tokens} & \multicolumn{3}{c|}{50  tokens} \\ \hline

         & CC & Pile & CC & Pile & Pile-C \\ \hline

        \makecell{Straight Dec} & 2.47 & 2.67 & 0.37 & 0.52 & 0.43  \\ \hline

        \makecell{U-Shape Dec} &  8.86 & 9.13 & 1.82 & 1.96 & 1.87 \\ \hline
         
        \makecell{Inverted-U- \\ Shaped Dec} & 6.25 & 6.35 & 1.90 & 2.09 & 1.97 \\ \hline
        
        \makecell{Straight Inc} & 31.85 & 20.83 & 55.37 & 54.86 & 55.89  \\ \hline

        \makecell{U-Shaped Inc} & 18.86 & 14.51 & 11.63 & 11.26 & 10.09 \\ \hline

        \makecell{Inverted-U- \\ Shaped Inc} & 41.45 & 33.14 & 32.68 & 33.02 & 33.76 \\ \hline
        
    \end{tabular}
    \label{tab:prefix-trends}
    \vspace{-0.1in}
\end{table}

\subsection{Effect of token position}
\begin{figure}
    \centering
    \includegraphics[trim={0.11in 0 0 0},clip]{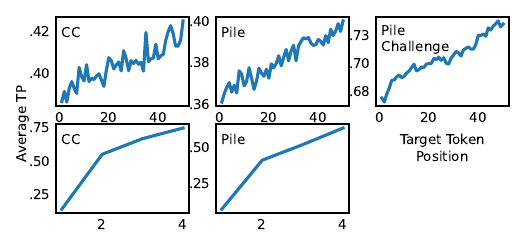}
    \vspace{-0.25in}
    \caption{Average Token Probability (TP) of a given token in the target suffix as a function of the position of the token.}
    \label{fig:tok-pos}
    \vspace{-0.25in}
\end{figure}

The first known work on text extraction~\cite{carlini2021extracting} suggests logit normalization through Softmax with a \emph{decaying} temperature to improve extraction rates. A decaying temperature flattens the token distribution for the earlier tokens while making the distribution more pronounced for later tokens. This has the effect of allowing the model to choose from a wide range of tokens at the beginning, while narrowing down the choices for later tokens. \cite{carlini2021extracting} reason that ``this gives a sufficient amount of time for the model to explore a diverse set of prefixes while also allowing it to follow a high-confidence path that it finds''. This implies that it should be easier to correctly extract later tokens, since the model does not need to explore many tokens in the latter parts of the suffix. Although this intuition has been upheld in other works~\cite{yu2023bag}, we can also use Token Probabilities to empirically verify whether it is easier to extract later tokens. For this, we plot the average Token Probability (TP; Equation~\ref{eq:tlp}) over all suffixes in our dataset as a function of each token's position. We plot our results in Figure~\ref{fig:tok-pos}. 

\paragraph{Discussion} Figure~\ref{fig:tok-pos} shows that earlier tokens in the target suffix are generally harder to extract than later ones, although by different margins. The last token of the Named Entities is on average $5.23-10.12$X easier to extract than the first token. However, the corresponding numbers are only $1.10-1.11$X for $50$-token sequences. This is despite the fact that the longer sequences get even more specific prompts for the final token, as they have the preceding $49$ tokens of the suffix in addition to the $50$-token prefix. Thus, the $50^{th}$ token of the $50$-token suffix receives a total of $50 + 49 = 99$ tokens of context. However, the final token for each Named Entity only gets $50 + 3 = 53$ tokens of context and yet are, on average, an order of magnitude easier to extract. This could be explained by LLMs' tendencies for stronger memorization of out-of-distribution text. Unlike most random $50$-token strings, Named Entities are not natural English, and thus might be easier for the LLMs to memorize. 

\section{Related Work}
\label{sec:rel}
\cite{carlini2019secret} was one of the first works to examine memorization in LLMs, and did so by introducing canaries during training and measuring their perplexity compared to random data during inference. More recently, \cite{duan2024membership,shi2023detecting} also explore such Membership Inference Attacks on LLMs. \cite{carlini2021extracting,carlini2022quantifying} measured leakage of training data and examined memorization scaling laws in LLMs through deterministic generation schemes like greedy decoding. \cite{yu2023bag} provide new methods to be able to increase leakage of training data from LLMs. In addition to deterministic generation, they also explore the more commonly deployed randomized generation schemes like top-$k$ and top-$p$, however their analysis is not at the sequence level. Concurrent work \cite{hayes2024measuring} also explores probabilistic metrics for quantifying memorization, However, they use population averages, unlike this work. \cite{lukas2023analyzing} study leakage of PII in LLMs. \cite{kassem2024alpaca} leverage one LLM for prompt-optimization to leak data from another LLM. \cite{biderman2024emergent} attempt to predict which sequences will be memorized by an LLM using lower-compute trial runs. \cite{kim2024propile} introduce the notion of sequence likelihood, the metric that we use for our analysis. \cite{tirumala2022memorization} explore scaling laws and examine memorization dynamics throughout training. \cite{lu2024scaling} study scaling laws for ``fact memorization'' within LLMs. \cite{huang2022large} study trends in both memorization, and \emph{association} of PII, where the latter involves leakage through prompts that are not verbatim training sequences.
\cite{feldman2020does} introduces the idea of the ``long-tail", where they show that models have a tendency to memorize rare, out-of-distribution samples. \cite{maini2023can} explore whether memorization can be localized to certain parts of the model.  \cite{ippolito2023preventing} argue that focusing solely on verbatim memorization provides an incomplete picture of privacy risks. \cite{hartmann2023sok} provides a systematic overview of the work done in the memorization space.

\section{Conclusion}
\label{sec:conclusion}
In this work, we first show how predominant metrics such as extraction rate underestimate the threat of training data leakage by up to $2.14$X. We also show how measuring an averaged extraction rate can often lead to conclusions different from a finer-grained sequence-level analysis. We do so by examining the effects of model size, prefix length, decoding schemes, and token position at the level of individual sequences. The insights gained
from our analysis show that it is important to look at leakage of training data on a per-sequence basis rather than merely averaged quantities. 

\section{Impact Statement}
The aim of this work is to better analyze memorization of training data in Large Language Models, which can have use in quantifying the risk of leaking privacy-sensitive data or copyrighted content.

\bibliography{main}
\bibliographystyle{icml2025}

\appendix
\section{Bounding Approx n-ISP}
\label{app:isp}



Equation~\ref{eq:inex-approx} and Figure~\ref{fig:islp-approx} show how we can approximate $n$-ISP by just iterating over the higher probability sequences in $\mathcal{E}(t_{j+1:j+m},n)$. Here we describe how to compute a loose upper bound on the error term $\varepsilon$ in Equation~\ref{eq:inex-approx}. 
Simply put, a theoretical upper bound on $\varepsilon$ could be obtained by summing the probabilities of the ignored tokens in each step of the suffix generation. 
To explain this further, let us assume that we have a model with a vocabulary of $32,000$ tokens indicated by $\mathcal{V}_{32K}$, and we want to compute $1$-ISP for a given target sequence of $4$-tokens \{ $t_{51}$, $t_{52}$, $t_{53}$, $t_{54}$ \} when prompted by a prefix $t_{1:50}$. First, we add up the probabilities of all strings where the first token is an incorrect match i.e., $\sum_{t \in \mathcal{V}_{32K} - t_{51}} P(t | t_{1:50})$. To do so, we prompt the model with the sequence prefix, using which it produces a probability distribution over the first (of four) tokens. Let us say, in order to approximate the probability, we only continue iteration with the top-$10$ tokens in this distribution that are not an exact match with the ground-truth token $t_{51}$. This means that we are approximating $\sum_{t \in \mathcal{V}_{32K} - t_{51}} P(t | t_{1:50})$ with $\sum_{t \in \mathcal{V}_{10}} P(t | t_{1:50})$, which means that instead of iterating over $\mathcal{V}_{32K} - t_{51}$ tokens, we only iterate over $\mathcal{V}_{10}$ tokens. In order to compute an upper bound, we assume that had the model been prompted with each of the $31,990$ ignored tokens, it would have produced the remaining $3$ tokens in the target sequence with a probability of $1$. Hence, the error term for this iteration would simply be $\sum_{t \in \mathcal{V}_{31990}} P(t | t_{1:50})$. Since $1$-ISP is the probability that \emph{any} one token is incorrect, this process is repeated for tokens $t_{52}$, $t_{53}$, and $t_{54}$ being incorrect. The error terms from each of them are summed together to give one final error term for $1$-ISP for the given sequence.

Since we assume the model provides perfect predictions for the ignored tokens, this gives us a worst-case upper bound on $\varepsilon$, which although likely to be much higher than the true probability, gives us a strict cap on how large the risk of leakage in that particular scenario will be. 

\section{Examples of Sequences}
\label{app:model-size-ex}
Table~\ref{tab:model-size-seq} shows examples of sequences that are easier to extract with smaller models. We note that when extracting Named Entities, both natural looking and unnatural text are easy to extract with small models. The first two rows of Table~\ref{tab:model-size-seq} show examples of both these types of sequences. However, for longer, $50$-token suffixes, most of the text that was easiest to extract with smaller models was unnatural looking. Row $3$ provides an example of such a sequence.

\begin{table*}[]
    \centering
    \begin{tabular}{|c|c|c|}
    \hline
    Prefix & Suffix & Model Size (ESP) \\ \hline
    \makecell{\dots limit Iran’s enrichment of uranium, a potential  \\ pathway to the development of \\ nuclear weapons. A German diplomatic source \\ confirmed that Foreign Minister} &  Heiko Maas & $125$M ($0.46$) \\ \hline
    \makecell{\dots  toolsVersion=``5053" systemVersion=``\\13E28" targetRuntime=``iOS.} & CocoaTouch  & $125$M ($0.0076$) \\ \hline
    \makecell{\dots 
  (int)(long)\&((struct stringpool\_t *)0} & \makecell{\dots )-\>stringpool\_str32, \dots \\ stringpool\_str525 \dots 0)-\> \dots} & $350$M ($1.18e^{-08}$) \\ \hline
  
    \end{tabular}
    \caption{Examples of sequences that are easier to extract with smaller models. The first two columns contain the truncated prefixes and suffixes, where ``\dots''~are used to show the truncation. Column $3$ shows which model had the highest probability of extracting the suffix, with the ESP in parentheses.}
    \label{tab:model-size-seq}
\end{table*}










\end{document}